\documentclass[11pt]{article}
\usepackage[margin=1in]{geometry}
\usepackage{amsmath,amssymb,amsthm}
\usepackage{booktabs}
\usepackage{tabularx}
\usepackage{tikz}
\usepackage{microtype}
\usepackage{enumitem}
\usepackage[hidelinks]{hyperref}
\usetikzlibrary{arrows.meta,positioning,shapes.geometric,calc,fit,backgrounds}

\newtheorem{definition}{Definition}
\newtheorem{proposition}{Proposition}
\newtheorem{remark}{Remark}

\newcommand{\nEVI}{\mathrm{nEVI}}
\newcommand{\Lclin}{L_{\mathrm{clin}}}
\newcommand{\Rset}{\mathcal{R}}

\title{\bf From Few-Shot Segmentation to Clinician-in-the-Loop \\ Medical Image Analysis:\\[2pt]
\large Decidable Self-Assessment as the Precondition for Interaction and Adaptation}
\author{Yazhou Zhu\\ \texttt{rangerzhuyazhou@gmail.com}}
\date{}

\begin{document}
\maketitle

\begin{abstract}
\noindent Few-shot medical image segmentation (FSMIS) seeks to delineate unseen structures from a small
support set, but its standard formulation fixes task-defining evidence before inference. This assumption is
fragile when query cases exhibit acquisition shift, atypical pathology, ambiguous boundaries, or poor image
quality. A natural response is to add clinician interaction and rapid adaptation. We argue that this response,
stated at that level, is incomplete: promptable interfaces and test-time adaptation are already available, so
the binding constraint is not the ability to ask or the ability to change, but the ability to \emph{decide}
whether asking or changing is warranted for the case at hand. Both capabilities depend on confidence
estimates that are calibrated on typical data and degrade precisely in the rare regime that motivates them.
This Perspective therefore reframes FSMIS as a three-layer problem. The base layer is decidable
self-assessment: a system must separate errors that a bounded intervention can repair from errors that no
admissible intervention can reach. We formalize this through the \emph{correctable set}, the family of
segmentations reachable under a bounded update operator and the remaining interaction budget, and show
that it supplies a decision rule for accept, query, and defer, rather than a cost comparison alone. The middle
layer is selective interaction: with a static support budget $K$ and a distinct interaction budget $B$, queries
are chosen by response-conditioned net expected value of information, and the default action is not to ask.
The upper layer is bounded adaptation, where the operative property is reversibility rather than speed. We
further promote cross-case experience from an optional extension to an explicit second memory layer, and
argue that what transfers across rare cases is not a mask prior but a correction prior over failure modes,
gated by cross-reader reproducibility. We restate the research agenda as six hypotheses with an explicit
dependency order, in which the value of querying is conditional on the validity of multi-level risk
estimation, and specify a minimal pilot that can falsify the foundational claim before any clinician study.
The central claim is not that interaction resolves domain shift, but that scarce expert attention should be
spent only where a bounded intervention is expected to reach a clinically better outcome.
\end{abstract}


\section{Introduction}

Few-shot medical image segmentation is motivated by a simple constraint: expert masks are expensive,
whereas new anatomical structures, pathologies, scanners, and imaging protocols appear continually. The
canonical answer conditions a segmentation model on a small set of labeled support images and infers the
mask for a query image. This framing has generated substantial progress, but it implicitly assumes that the
initial support set contains enough evidence to resolve the query. That assumption is weakest precisely in
the cases for which adaptable medical imaging systems are most valuable: rare disease, atypical anatomy,
postoperative change, severe artifact, unfamiliar acquisition, or a genuinely ambiguous boundary.

A widely shared intuition is that the next step is therefore twofold: let the model interact with the clinician,
and let it adapt quickly from what the clinician says. We agree with the direction and disagree with the level
at which it is usually stated. Interaction and adaptation are \emph{execution} capabilities, and both are, in
isolation, largely solved. Promptable segmentation supplies a general interaction interface
\cite{kirillov2023sam,wong2024scribbleprompt}; test-time adaptation, adapters, and prompt optimization
supply mechanisms for fast change \cite{karani2021tta,wu2025medsamadapter,xu2026oaims}. Section
\ref{sec:gap} documents that human correction, context accumulation, support selection, deferral, and
preference alignment are all individually established. If the missing ingredient were the ability to ask or the
ability to change, the problem would already be closed.

What is missing is the \emph{judgment} that governs both: knowing when to ask, what to ask, and when not
to change. These three judgments share a single dependency, namely an estimate of residual risk for the
present case, and that estimate fails in exactly the regime under discussion. Modern networks are
overconfident \cite{guo2017calibration}, calibration degrades under distribution shift
\cite{ovadia2019trust}, and atypicality scores measure how unusual a case is rather than how likely it is to
fail \cite{hendrycks2017baseline}. An interaction policy built on such a signal will interrupt clinicians on
unusual but easy cases and release confident errors on familiar but hard ones. An adaptation rule built on it
will update on the wrong evidence and cannot detect that it has done so.

This Perspective consequently organizes clinician-in-the-loop few-shot analysis into three layers with a
strict dependency order.

\begin{description}[leftmargin=1.6em,itemsep=2pt]
\item[Layer 1: decidable self-assessment.] The system must distinguish an error that a bounded intervention
can repair, an error that no admissible intervention can reach, and an apparent uncertainty that does not
correspond to error at all. Without this layer, interaction degenerates into asking the clinician to keep
clicking, and adaptation degenerates into drift.
\item[Layer 2: selective interaction.] Given the first layer, the operative virtue of interaction is parsimony
rather than speed. Clinician attention is the most expensive resource in the system, and an unnecessary
interruption can cost more clinically than a minor boundary deviation. The default action is to \emph{not}
ask, and a query must justify itself.
\item[Layer 3: bounded adaptation.] Given the first two layers, the operative virtue of adaptation is
reversibility rather than speed. One-sample updates from possibly erroneous feedback, whose effects can
propagate beyond the edited region, are a structural hazard; ``fast'' is not the right objective.
\end{description}

Stated as a single sentence, the position of this paper is that the next step for few-shot medical image
analysis is not to extract a stronger representation from fewer examples, but to give a system a decidable
account of its own failure, so that it calls a human only at the few moments when a human can actually
help, and changes itself only within the range that evidence supports.

\paragraph{Contributions.} First, we formalize static task evidence and sequential clinician feedback as
separate resources, $K$ and $B$, which prevents ordinary interactive segmentation from being relabeled as
few-shot learning. Second, we introduce the correctable set and an associated decidability condition, which
converts accept/query/defer from a cost comparison into a partially decidable test and supplies a formal
definition of correctable versus non-correctable failure. Third, we narrow the novelty claim explicitly:
what remains untested is response-conditioned selection \emph{among heterogeneous feedback channels}
coupled to release decisions, not interaction, deferral, or adaptation as such. Fourth, we promote cross-case
experience to an explicit second memory layer and argue that the transferable object is a correction prior
over failure modes rather than a mask prior over rare diseases, with cross-reader reproducibility as the write
gate. Fifth, we address the cold-start problem of an unknown response model through a pessimistic value of
information. Sixth, we restate the agenda as six hypotheses with an explicit dependency order and specify a
minimal pilot that can falsify the foundational hypothesis before any clinician is recruited. Segmentation is
the motivating and formally developed case; extension to reconstruction, detection, or diagnosis is outside
the present claims. The synthesis is a selective Perspective rather than a systematic review, and literature
claims are restricted to the cited design space.

\section{Problem Foundations: Scarcity, Shift, and Ambiguity}

\subsection{Three forms of scarcity}

The motivation for FSMIS is normally described as label scarcity. In clinical edge cases this is only one of
three interacting constraints. \emph{Data scarcity} reflects the specialist labor required for dense masks.
\emph{Coverage scarcity} arises because unusual protocols, rare pathology, atypical anatomy, and severe
artifacts are intrinsically uncommon. \emph{Attention scarcity} reflects the fact that a clinician cannot
inspect and correct every model output. Improving average Dice under a fixed one-shot protocol addresses
part of the first constraint and leaves the other two implicit. The three are not independent: coverage
scarcity is what makes a case hard, and attention scarcity is what makes it costly to resolve, so a policy that
optimizes one in isolation will typically spend the other.

Acquisition, representation, and interpretation are also coupled. MRI pulse sequences, field strengths,
reconstruction pipelines, and vendor-specific choices alter contrast and texture; CT phase, dose,
reconstruction kernel, and hardware alter noise and edge appearance. Pathology can simultaneously change
target morphology and the clinical meaning of an uncertain boundary. Consequently the same apparent
segmentation failure may arise from missing task evidence, a shifted feature geometry, an ambiguous
reference standard, or a model limitation. Table \ref{tab:difficulty} separates these sources; a clinically
useful controller must not treat them as interchangeable, and in particular must not collapse them into a
single out-of-distribution label.

\begin{table}[t]
\centering
\small
\caption{Distinct sources of difficulty and their consequences for a clinician-in-the-loop policy. The
categories can co-occur but should not be collapsed into a single OOD label. The final column anticipates
Section \ref{sec:correctable}: only the first two rows are generally correctable by bounded intervention.}
\label{tab:difficulty}
\begin{tabularx}{\textwidth}{@{}l X X X@{}}
\toprule
\textbf{Source} & \textbf{Operational meaning} & \textbf{Typical evidence} & \textbf{Policy implication} \\
\midrule
Label-space novelty & The target structure or lesion was not represented among meta-training classes. &
Small labeled support set; semantic prompt; anatomical relation. & Improve task specification; query
semantic identity when the support set is insufficient. \\
\addlinespace
Domain shift & Meta-training and deployment differ in scanner, protocol, site, modality, or appearance. &
Domain-robust features; acquisition metadata; target support examples. & Reweight or adapt representation
while monitoring transfer and retention. \\
\addlinespace
Case-level OOD & The individual case lies outside the validated deployment envelope. & Atypicality score,
predicted mask quality, provenance, clinical consequence. & Use as a routing signal; do not equate
atypicality with failure probability. \\
\addlinespace
Clinical ambiguity & More than one contour can remain plausible even under matched acquisition. &
Multiple readers, provenance, task intent, downstream tolerance. & Represent disagreement, ask a
task-specific question, or defer to adjudication. \\
\bottomrule
\end{tabularx}
\end{table}

\subsection{Two independent supervision budgets}
\label{sec:budgets}

Let $K$ denote the number of static support image-mask pairs that initially specify an unseen class or task.
Let $B \in \mathbb{R}^m_+$ denote a vector of hard interaction limits available \emph{after} observing the
query, such as maximum elapsed time, number of interruptions, or corrected area; $b(a) \in \mathbb{R}^m_+$
measures the same resources for action $a$, and inequalities are componentwise. Each experiment should
prespecify the active components and units of $B$. A separate soft cost $c(a)$ represents burden not already
hard-constrained, with accounting rules that prevent double counting.

``Few-shot'' refers to small $K$; the extension proposed here asks how $B$ should be spent. This separation
does two kinds of work. It prevents ordinary interactive segmentation from being relabeled as few-shot
learning, and it prevents a gradually accumulated target-domain dataset from being mistaken for a bounded
per-case episode. Because the distinction is load-bearing for every subsequent claim, it is stated here rather
than deferred: a method that consumes unbounded $B$ has not solved a few-shot problem, whatever its $K$.

\subsection{Central thesis}

Reliable segmentation for rare, ambiguous, and shifted cases should be formulated as a sequential decision
problem in which the first decision is epistemic. A model uses a small support set as initial task evidence,
estimates not only what remains uncertain but whether that uncertainty is \emph{reachable} by an
admissible intervention, and spends limited clinician attention only when the expected reduction in clinical
risk justifies the cost. Reliability must emerge from externally validated decision rules,
response-conditioned querying, bounded and reversible adaptation, and principled deferral, not from model
confidence or model scale alone.

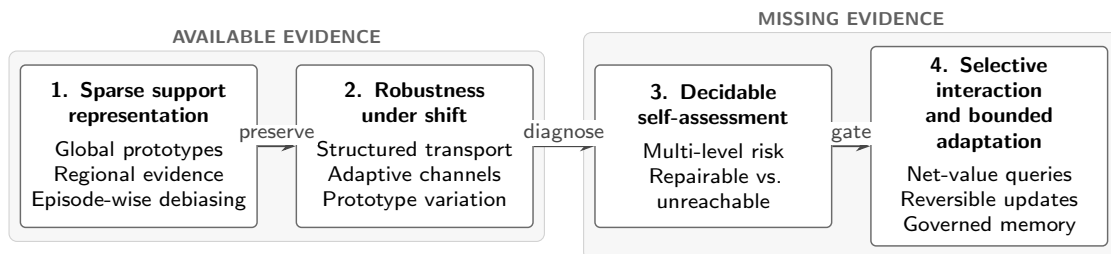
\begin{figure}[t]
\centering
\begin{tikzpicture}[
  box/.style={draw=black!58,line width=0.55pt,rounded corners=2pt,fill=white,
    align=center,inner sep=4pt,minimum height=2.15cm,text width=2.85cm,
    font=\sffamily\scriptsize},
  group/.style={fill=black!3,draw=black!18,line width=0.4pt,rounded corners=3pt,
    inner xsep=4pt,inner ysep=5pt},
  grouptitle/.style={font=\sffamily\fontsize{6.8}{7.5}\selectfont\bfseries,
    text=black!62,fill=white,inner xsep=2pt,inner ysep=0.5pt},
  ar/.style={-{Stealth[length=1.9mm,width=1.35mm]},draw=black!72,line width=0.7pt},
  lbl/.style={font=\sffamily\scriptsize,text=black!72,fill=white,inner sep=1.2pt}
]
\node[box] (a) {\textbf{1. Sparse support\\representation}\\[3pt]
  Global prototypes\\Regional evidence\\Episode-wise debiasing};
\node[box,right=5mm of a] (b) {\textbf{2. Robustness\\under shift}\\[3pt]
  Structured transport\\Adaptive channels\\Prototype variation};
\node[box,right=8mm of b] (c) {\textbf{3. Decidable\\self-assessment}\\[3pt]
  Multi-level risk\\Repairable vs. unreachable};
\node[box,right=5mm of c] (d) {\textbf{4. Selective interaction\\and bounded adaptation}\\[3pt]
  Net-value queries\\Reversible updates\\Governed memory};
\draw[ar] (a)--(b) node[midway,above,lbl]{preserve};
\draw[ar] (b)--(c) node[midway,above,lbl]{diagnose};
\draw[ar] (c)--(d) node[midway,above,lbl]{gate};
\begin{scope}[on background layer]
\node[group,fit=(a)(b)] (available) {};
\node[group,fit=(c)(d)] (missing) {};
\end{scope}
\node[grouptitle,anchor=south] at ([yshift=2pt]available.north) {AVAILABLE EVIDENCE};
\node[grouptitle,anchor=south] at ([yshift=2pt]missing.north) {MISSING EVIDENCE};
\end{tikzpicture}
\caption{Conceptual trajectory. The first two stages improve what a small support set retains; the third is
the precondition introduced here, and the fourth is what it licenses. The ordering is logical rather than a
claim of strict historical succession.}
\label{fig:trajectory}
\end{figure}

\section{From Fixed Support to Cross-Domain Robustness}

To motivate a decision-theoretic framework, limitations in how sparse evidence is \emph{represented} must
first be distinguished from limitations caused by evidence that is \emph{absent}.

\subsection{The canonical few-shot assumption}

One-shot semantic segmentation established that a single annotated example could specify a new
dense-prediction task at inference time \cite{shaban2017oneshot}. Prototypical Networks supplied a
metric-learning principle in which each class is represented by the center of its support embeddings
\cite{snell2017protonet}; PANet translated this principle to dense prediction through masked prototypes and
support-query alignment \cite{wang2019panet}. In medical imaging, volumetric task conditioning,
self-supervised representation learning, and anomaly-inspired foreground-background modeling have
relaxed this idealized geometry \cite{roy2020squeeze,ouyang2022ssl,hansen2022anomaly}. The common
pipeline nevertheless remains fixed: encode the support set and query, compress the labeled foreground into
a task representation, and transfer that representation to an unseen class.

Medical images expose the statistical weakness of this compression. Foreground and background are highly
imbalanced, labeled base structures can be scarce, and a structure rarely forms one compact cluster across
slices, patients, modalities, and disease states. The support mask is therefore not merely small; it is an
incomplete and potentially biased description of a heterogeneous target.

\subsection{Representing heterogeneous support evidence}

A sequence of within-domain studies can be read as complementary relaxations of the single-prototype
assumption rather than as a list of architectures. Three moves recur. \emph{Regional decomposition}
replaces one global prototype with several representative subregions, on the premise that a structure is a
mixture rather than a point \cite{zhu2023rept,zhu2024pami}. \emph{Relation-aware refinement} lets support
and query features rectify one another instead of treating the support summary as fixed
\cite{zhu2023rept}. \emph{Episode-conditioned filtering} challenges masked average pooling directly by
removing episode-specific foreground features before prototype construction \cite{zhu2024debiased}.
Comparable strategies appear outside this line in the broader domain-generalization literature, where
feature-level regularization and semantic-alignment objectives play the same role of preventing a single
summary statistic from absorbing nuisance variation \cite{dou2019masf,zhou2023dgsurvey}.

The theoretical continuity is more important than the individual method: what must a few labeled images
retain in order to represent a heterogeneous anatomical structure? All three moves imply that elements of a
support mask have unequal relevance and that relevance depends on the target case. None of them can create
relevance where the support set contains none.

\subsection{Cross-domain FSMIS}

CD-FSMIS introduces a second generalization axis: source-trained meta-knowledge must transfer to both
unseen classes and an unfamiliar target domain \cite{lei2022cdfss}. Even when test support and query
images originate from the same target domain, the encoder and similarity function learned on source data
may provide a poorly aligned coordinate system; an acquisition mismatch between test support and query
compounds the problem.

Recent work addresses complementary failure points. RobustEMD replaces point-wise prototype comparison
with structured transport between decomposed support and query features, with texture- and structure-aware
weighting suppressing domain-sensitive nodes \cite{zhu2025robustemd}. Dynamic Semantic Matching
performs support-query reweighting, dynamically selects domain-robust channels, and estimates semantic
centers from dual perspectives \cite{zhu2025dsm}. Adversarial Prototypical Perturbation constructs
gradient-derived perturbations from intra- and inter-class variation objectives and combines them with
local-imbalance-aware whitening \cite{zhu2026app}. MAUP uses DINOv2 features and a frozen Segment
Anything Model with multi-center prompt generation, uncertainty-aware prompt selection, and adaptive
prompt optimization without parameter training \cite{zhu2025maup}. These branches are complementary:
one improves the representation of sparse task evidence, another improves matching under shift, and the
foundation-model branch supplies a promptable interface.

\subsection{What better representation cannot recover}

Better support representations and domain-robust matching preserve available evidence. They cannot
recover case-specific anatomy or clinical intent that the support set never contained, and none of them
establishes whether a missing item of evidence is worth requesting from a clinician. This is the boundary at
which representation research ends and decision research begins, and it is the boundary this Perspective is
about.

\section{Adjacent Paradigms and the Integration Gap}
\label{sec:gap}

\subsection{Interaction is necessary but not sufficient}

Interactive medical segmentation has established that points, scribbles, boxes, and image-specific
optimization can correct a current output
\cite{wang2018ift,wang2019deepigeos,luo2021mideepseg,koohbanani2020nuclick}. ScribblePrompt
demonstrated generalization to unseen biomedical tasks and improved annotation efficiency relative to SAM
ViT-B in a study involving 16 academic-hospital imaging researchers who were shown target masks
\cite{wong2024scribbleprompt}. This supports known-target annotation efficiency, not independent
patient-level boundary judgment.

Interaction is also not universally reactive. PseudoClick predicts candidate next clicks automatically
\cite{liu2022pseudoclick}; MECCA combines action confidence with reinforcement learning to guide the
next interaction region \cite{shen2023mecca}; sequential-memory models encode the order of user
corrections \cite{mikhailov2024seqmem}. Interactive few-shot learning uses simulated point corrections for
regularized test-time optimization \cite{feng2021ifsl}, while IFSS-Net propagates expert-seeded information
in volumetric ultrasound \cite{alchanti2021ifssnet}. Correction-driven adaptation across images and domain
change is likewise established \cite{kontogianni2020continuous}.

More recent systems occupy additional parts of the design space: clinician-corrected predictions for
cross-center test-time adaptation \cite{hu2024clinicianpreferred}; multiple plausible masks for iterative
refinement \cite{zhu2024meduhip}; image ordering as context grows \cite{torpey2026ordering}; support
selection and in-context failure detection \cite{gehad2026context}; pixel- or region-level deferral to multiple
experts \cite{tian2026deferredseg}; online adaptation from user-refined masks under distribution shift
\cite{xu2026oaims}; and uncertainty-guided candidate selection for preference alignment
\cite{zhao2026uair}. Their evidence levels differ: several are preprints, one is a workshop contribution, and
others are peer-reviewed conference or journal studies.

We draw the conclusion explicitly, because it constrains what this Perspective may claim. Human
correction, adaptation, context accumulation, support selection, deferral, and preference alignment are
\emph{not novel in isolation}. Any position paper whose thesis is ``few-shot systems should interact and
adapt'' is therefore restating existing capability. The open problem must lie elsewhere.

\subsection{Active learning, in-context learning, and deferral}

Pool-based active learning asks which samples or regions should be labeled to improve a future global
model \cite{settles2009survey}. Information-theoretic and Bayesian acquisition formalize informativeness
through expected information gain or epistemic uncertainty \cite{mackay1992ibof,gal2017dbal}; medical
systems combine these signals with representativeness, consistency, diversity, or simplified interactive
labels \cite{yang2017suggestive,li2023halia}. Interactive segmentation usually targets the current mask,
whereas in-context segmentation uses completed examples to specify a task without gradient-based
retraining: UniverSeg demonstrates broad in-context medical segmentation \cite{butoi2023universeg},
MultiverSeg accumulates completed examples as context \cite{wong2025multiverseg}, and the ordering of
these examples can alter both accuracy and effort \cite{torpey2026ordering}. Deferral has an established
decision-theoretic basis in learning-to-defer \cite{madras2018predict,mozannar2023whopredict}, and
classical information value theory asks whether an observation changes a decision \cite{howard1966ivt}.

\subsection{The narrowed claim}

Given the preceding two subsections, we state the residual gap in the narrowest form that survives the
literature. What remains insufficiently tested is not another interaction primitive, and not the general
proposition that interaction helps. It is a controlled integration of three specific elements:

\begin{enumerate}[leftmargin=1.8em,itemsep=2pt]
\item a \emph{decidable} account of which current-case errors a bounded intervention can reach, evaluated
under explicit few-shot, cross-domain, and case-level OOD conditions;
\item response-conditioned selection \emph{among heterogeneous feedback channels} whose bandwidth and
semantics differ, coupled to accept/query/defer release decisions rather than to mask refinement alone;
\item joint evaluation of risk, effort, adaptation stability, and escalation on the same trajectory.
\end{enumerate}

Net expected value of information is itself classical \cite{howard1966ivt} and standard in active learning;
we claim no novelty for it. The claim is the coupling of channel-level value estimation to a release decision
that is gated by a reachability test.

\begin{table}[t]
\centering
\small
\caption{Primary objectives of related paradigms. The rows are not mutually exclusive, and existing methods
already bridge several categories; the final row states what the present framework adds rather than what it
invents.}
\label{tab:paradigms}
\begin{tabularx}{\textwidth}{@{}l X X X@{}}
\toprule
\textbf{Paradigm} & \textbf{When supervision arrives} & \textbf{What supervision changes} & \textbf{Central limitation or advance} \\
\midrule
Few-shot segmentation & Fixed support before inference & Task representation or prototypes & No mechanism to acquire missing case evidence. \\
\addlinespace
Interactive segmentation & Corrections during inference & Current mask; sometimes current image representation & Optimizes refinement rather than joint risk and release. \\
\addlinespace
Pool-based active learning & Selected labels during development & Future global model & Does not manage current-case release risk. \\
\addlinespace
In-context segmentation & Examples accumulate as context & Task specification without weight updates & Selection, calibration, and deferral remain incompletely coupled. \\
\addlinespace
Learning to defer & At routing time & Who decides & Routes on predicted error, not on predicted repairability. \\
\addlinespace
Proposed framework & Model-initiated multimodal feedback under budget, gated by reachability & Case, task, and domain representation, plus governed memory & Intended to jointly optimize and evaluate error, effort, escalation, and retention. \\
\bottomrule
\end{tabularx}
\end{table}

\section{Decision-Theoretic Formulation}

\subsection{Episode state, actions, and bounded update}

An episode begins with a query image or volume $x$, $K$ support image-mask pairs $S_K$, an initial model
$f_0$, and interaction budget $B$. At step $t$ the state
\begin{equation}
s_t = (x, S_K, h_t, f_t, B_t, u_t)
\end{equation}
contains the interaction history $h_t$, current model or episode representation $f_t$, remaining budget
$B_t$, and a hierarchy of local, structural, and case-level risk evidence $u_t$. The controller chooses
$d_t \in \{A, Q, D\}$: accept the current mask, query the clinician through action $a_t \in \mathcal{A}_Q$, or
defer to full review. A query action jointly specifies a \emph{location}, a \emph{modality}, and a
\emph{question}. The response $z_t$ may include disagreement, correction error, no response, or latency. A
bounded transition operator $\mathcal{T}(s_t, a_t, z_t)$ can update prompts, prototypes, memory, latent
features, or a restricted parameter subset. The episode terminates at stopping time $\tau$.

Acceptance and deferral have different consequences. Acceptance incurs clinically weighted loss from a
released mask. Deferral incurs full-review time and the residual loss of the expert or joint result; it is
neither free nor perfect. For feasible policies $\Pi(B) = \{\pi : \sum_{t<\tau} b(a_t) \preceq B\}$, where
$\preceq$ is componentwise, consider
\begin{equation}
\label{eq:objective}
\begin{split}
\pi^\star = \arg\min_{\pi \in \Pi(B)} \mathbb{E}_\pi \Big[
&\ \mathbb{I}(d_\tau = A)\,\Lclin(\hat{y}_\tau, y)
+ \mathbb{I}(d_\tau = D)\{C_D + \Lclin(y^D_\tau, y)\} \\
&+ \lambda_C \sum_{t<\tau} c(a_t)
+ \lambda_R \big[L_\Rset(f_\tau) - L_\Rset(f_0) - \varepsilon_R\big]_+ \Big].
\end{split}
\end{equation}
Here $b(a)$ records prespecified hard resource use, whereas $c(a)$ grades residual burden not already
represented in the active components of $b(a)$; the cost ledger and normalization are fixed before evaluation
to prevent double counting. The final term penalizes degradation beyond tolerance $\varepsilon_R$ on an
immutable protected reference set $\Rset$. Clinical losses must be elicited for a defined downstream
decision: a missed lesion, a critical boundary violation, and a harmless surface discrepancy cannot share an
arbitrary common weight.

\subsection{The correctable set and the decidability condition}
\label{sec:correctable}

Equation \eqref{eq:objective} is a cost comparison. On its own it does not say whether a query \emph{can}
help, only whether it would be cheap. This matters because of an information-theoretic asymmetry that the
interaction literature rarely states: a click carries a few bits, whereas domain shift induces a
high-dimensional mismatch in the representation. Low-bandwidth input cannot inject high-dimensional
information. What it can do is \emph{collapse ambiguity} among hypotheses the model already entertains.
Interaction therefore helps only when an acceptable solution lies within the range the system can reach.

We make this precise. Let $\mathcal{A}_Q(B_t)=\{a \in \mathcal{A}_Q : b(a) \preceq B_t\}$ be the admissible
query actions under the remaining budget, and let $\mathcal{Z}(a)$ be the support of plausible responses to
$a$.

\begin{definition}[Reachable set]
The one-step reachable set at state $s_t$ is
\[
\mathcal{M}_1(s_t) = \{\hat{y}(\mathcal{T}(s_t,a,z)) : a \in \mathcal{A}_Q(B_t),\; z \in \mathcal{Z}(a)\},
\]
and the budget-limited reachable set $\mathcal{M}(s_t)$ is its closure under admissible action sequences whose
cumulative cost satisfies $\sum b(a) \preceq B_t$.
\end{definition}

\begin{definition}[Correctable set]
Given a clinically elicited acceptability criterion $\mathcal{C}$ with tolerance $\eta$, the correctable set at
$s_t$ is $\mathcal{M}^{\mathcal{C}}(s_t) = \{\hat{y} \in \mathcal{M}(s_t) : \mathcal{C}(\hat{y}; \eta) = 1\}$.
The current case is \emph{correctable} at $s_t$ if $\mathcal{M}^{\mathcal{C}}(s_t) \neq \emptyset$.
\end{definition}

Three consequences follow. First, correctability is defined relative to the update operator and the remaining
budget, not as a property of the image; enlarging $\mathcal{T}$ or $B$ can make a case correctable, which
is precisely what a safety argument must bound. Second, the qualitative distinction between
repairable and unreachable failure acquires a formal meaning rather than
resting on descriptive language. Third, the accept/query/defer rule gains a structural component:

\begin{proposition}[Reachability gate]
\label{prop:gate}
If $\mathcal{M}^{\mathcal{C}}(s_t) = \emptyset$, then for every $a \in \mathcal{A}_Q(B_t)$ the post-response
state cannot attain an acceptable mask, and any strictly positive soft cost $c(a) > 0$ implies
$\nEVI(a \mid s_t) \le 0$. The correct action is therefore $D$ (or $A$, if the current mask is already
acceptable), not $Q$.
\end{proposition}

The proposition is deliberately weak, and its usefulness lies in what it excludes rather than in what it
proves. In practice $\mathcal{M}^{\mathcal{C}}$ is not computable, because $\mathcal{C}$ depends on the
unobserved reference standard. It must therefore be replaced by an estimated reachability score
$\hat{\rho}(s_t) \approx \Pr[\mathcal{M}^{\mathcal{C}}(s_t) \neq \emptyset]$, trained on simulated and
recorded interaction traces where the post-hoc reachable outcome is known. Two properties make
$\hat{\rho}$ a more tractable estimation target than case-level accuracy itself. It is a function of the
model's own output geometry and the admissible operator, both of which are observable at inference time,
and it does not require identifying the correct contour, only whether an acceptable one is within range. We
regard the empirical behavior of $\hat{\rho}$ under shift as the single most informative experiment this
framework proposes; if $\hat{\rho}$ cannot be estimated better than chance at an external site, the
three-layer architecture fails at its base and the rest of the agenda is moot.

\begin{remark}
The reachability gate also clarifies the relation between this framework and deferral research. Learning to
defer typically routes on predicted model error \cite{madras2018predict,mozannar2023whopredict}. The gate
routes on predicted \emph{irreparability}, which is a different quantity: a case may be badly wrong and
trivially correctable, or nearly right and unreachable because the residual disagreement is irreducible
ambiguity rather than error.
\end{remark}

\subsection{Response-conditioned value of clinical feedback}

Uncertainty alone is not a reason to interrupt a clinician. A query is useful only if a plausible response is
expected to change the mask, lower residual risk, or determine that the case requires deferral. Define the
retention penalty $P_\Rset(s) = \lambda_R[L_\Rset(f_s) - L_\Rset(f_0) - \varepsilon_R]_+$. Let $J_A(s)$ and
$J_D(s)$ denote the expected costs of accepting and deferring, each including $P_\Rset(s)$, and let
$J_{\mathrm{stop}}(s) = \min\{J_A(s), J_D(s)\}$. The net one-step expected value of information is
\begin{equation}
\label{eq:nevi}
\nEVI(a \mid s_t) = J_{\mathrm{stop}}(s_t) - \mathbb{E}_{z \sim p(z \mid s_t, a)}\big[J_{\mathrm{stop}}(\mathcal{T}(s_t,a,z))\big] - \lambda_C c(a).
\end{equation}
A greedy controller selects
\begin{equation}
\label{eq:greedy}
a^g_t = \arg\max_{a \in \mathcal{A}_Q(B_t)} \nEVI(a \mid s_t)
\quad\text{subject to}\quad \hat{\rho}(s_t) \ge \rho_{\min},
\end{equation}
and queries only when the maximum is positive. The reachability constraint is stated explicitly in
\eqref{eq:greedy} because it is not implied by positivity of $\nEVI$ under a misspecified response model: a
response model that is optimistic about what a click achieves can manufacture positive value for an
unreachable case. The non-myopic form of the controller is given in Appendix \ref{app:bellman}; it matters
because one answer can alter the value of later questions, but it is not required to state the position of this
paper.

Channel semantics differ and should not be conflated. A click is inexpensive but low bandwidth; a boundary
correction conveys geometry; a reference case or short text may convey domain context or task intent. The
last two are hypotheses to be tested, not capabilities implied by click- or box-prompt models. Because
$J_A$ and $J_D$ contain $P_\Rset$, a query cannot acquire positive value merely by sacrificing protected
capabilities. Candidate updates must also be reversible: when a separately frozen monitor rejects a
post-response state, $\mathcal{T}$ returns the pre-update state and routes the case to another query or to
deferral. Equations \eqref{eq:objective}--\eqref{eq:greedy} are normative until their risk, response, cost,
and update models have been estimated for a defined clinical use case and separately safeguarded.

\subsection{Cold start: acting under an unknown response model}
\label{sec:coldstart}

Equation \eqref{eq:nevi} requires $p(z \mid s, a)$, the distribution of clinician responses. At deployment
onset this distribution is unknown, and it is exactly then that a miscalibrated policy does the most damage
to trust. We therefore treat the response model as an explicit staged object rather than an assumption.

Let $\mathcal{P}_a$ be an ambiguity set of response distributions consistent with prior evidence for action
$a$: for instance, all distributions within a divergence ball of a simulated-user model, or all mixtures of an
expert-agreement model and a no-response model with bounded weight. Define the pessimistic net value
\begin{equation}
\label{eq:pnevi}
\underline{\nEVI}(a \mid s_t) = J_{\mathrm{stop}}(s_t) - \sup_{p \in \mathcal{P}_a} \mathbb{E}_{z \sim p}\big[J_{\mathrm{stop}}(\mathcal{T}(s_t,a,z))\big] - \lambda_C c(a),
\end{equation}
and require $\underline{\nEVI}(a \mid s_t) > 0$ during the cold-start phase. Because $\underline{\nEVI} \le
\nEVI$, this is conservative by construction: it suppresses queries whose value depends on an optimistic
reading of how clinicians respond, at the cost of some missed opportunities. The phase transition should be
prespecified rather than discretionary. A workable rule is to shrink $\mathcal{P}_a$ toward the empirical
response distribution once a minimum number of observed responses per channel and per stratum is
reached, with the minimum fixed in advance and the shrinkage schedule reported; strata that never reach
the minimum remain permanently pessimistic. Sensitivity of the resulting policy to $\mathcal{P}_a$ should
be reported alongside its performance, since a policy whose behavior is invariant to the ambiguity set is not
actually using response information.

\subsection{Accept, query, or defer}

Querying is appropriate when risk is elevated, reachability is plausible, and a compact intervention is likely
to resolve the case. Deferral is appropriate when the case lies outside the validated adaptation envelope,
support evidence is inadequate, ambiguity is irreducible, reachability is low, or another interaction has low
expected value. Acceptance is permitted only by an empirical decision rule whose clinically defined
threshold is frozen after stratified external validation. Selective prediction provides the risk-coverage
perspective \cite{geifman2019selectivenet}; selective medical segmentation demonstrates why performance
on a confident subset differs from average accuracy \cite{ding2020selective}. Distribution-free risk control
supplies a language for stating what a frozen threshold does and does not guarantee
\cite{bates2021rcps}, and its exchangeability assumptions are precisely what external-site deployment
strains.

The clinician is not assumed to be an infallible oracle. Boundaries may be ambiguous, experts may disagree,
and a hurried correction may be wrong. Feedback should retain provenance, modality, timing, and
confidence. STAPLE models a probabilistic consensus and rater-specific performance under
conditional-independence assumptions \cite{warfield2004staple}; probabilistic segmentation, hierarchical
probabilistic models, and MedUHIP illustrate alternative ways to represent several plausible masks
\cite{kohl2018probunet,baumgartner2019phiseg,zhu2024meduhip}. The framework must distinguish
annotation error, systematic reader preference, irreducible ambiguity, and legitimate use-case-specific
contouring rather than collapsing them into one latent truth. Section \ref{sec:memory} makes this
distinction operational, because it determines what may be written to memory.

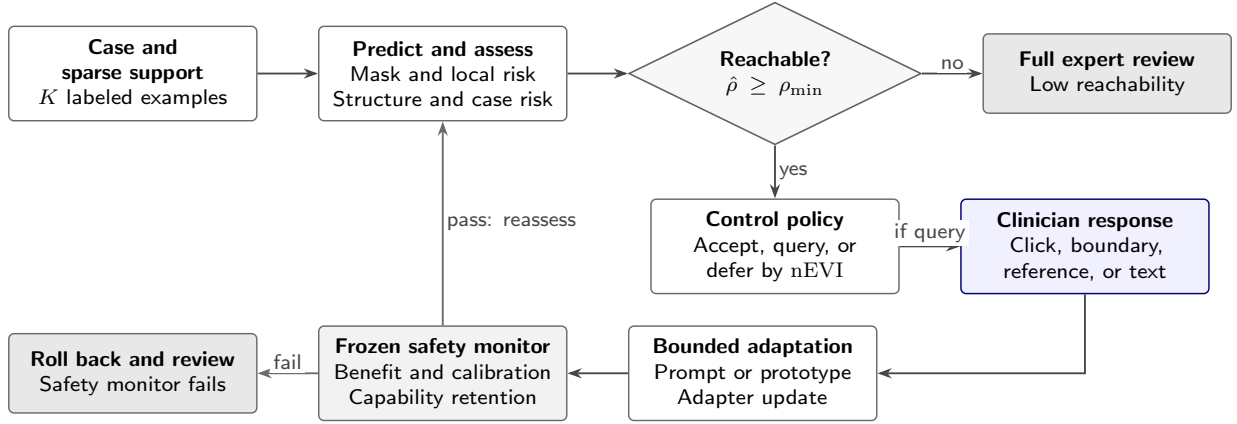
\begin{figure}[t]
\centering
\begin{tikzpicture}[
  b/.style={draw=black!58,line width=0.55pt,rounded corners=2pt,fill=white,
    align=center,font=\sffamily\scriptsize,inner sep=4pt,text width=3.0cm,
    minimum height=1.25cm},
  clinician/.style={b,fill=blue!6,draw=blue!45!black},
  monitor/.style={b,fill=black!5},
  terminal/.style={b,fill=black!9,minimum height=1.05cm},
  d/.style={draw=black!62,line width=0.6pt,diamond,aspect=2.05,fill=black!3,
    align=center,font=\sffamily\scriptsize,inner sep=1.5pt,text width=2.35cm},
  ar/.style={-{Stealth[length=1.9mm,width=1.35mm]},draw=black!72,line width=0.7pt},
  loop/.style={ar,draw=black!58},
  lbl/.style={font=\sffamily\scriptsize,text=black!72,fill=white,inner sep=1.2pt}
]
\node[b] (case) {\textbf{Case and sparse support}\\$K$ labeled examples};
\node[b,right=8mm of case] (pred) {\textbf{Predict and assess}\\Mask and local risk\\Structure and case risk};
\node[d,right=8mm of pred] (gate) {\textbf{Reachable?}\\$\hat\rho \ge \rho_{\min}$};
\node[terminal,right=8mm of gate] (defer) {\textbf{Full expert review}\\Low reachability};
\node[b,below=7mm of gate] (pol) {\textbf{Control policy}\\Accept, query, or defer by $\nEVI$};
\node[clinician,right=8mm of pol] (resp) {\textbf{Clinician response}\\Click, boundary, reference, or text};
\node[monitor,below=27mm of pred] (mon) {\textbf{Frozen safety monitor}\\Benefit and calibration\\Capability retention};
\node[b,right=8mm of mon] (adapt) {\textbf{Bounded adaptation}\\Prompt or prototype\\Adapter update};
\node[terminal,left=8mm of mon] (rollback) {\textbf{Roll back and review}\\Safety monitor fails};

\draw[ar] (case)--(pred);
\draw[ar] (pred)--(gate);
\draw[ar] (gate)--(defer) node[midway,above,lbl]{no};
\draw[ar] (gate)--(pol) node[midway,right,lbl]{yes};
\draw[ar] (pol)--(resp) node[midway,above,lbl]{if query};
\draw[ar] (resp.south)|-(adapt.east);
\draw[ar] (adapt)--(mon);
\draw[loop] (mon)--(rollback) node[midway,above,lbl]{fail};
\draw[loop] (mon.north)--(pred.south) node[midway,right,lbl]{pass: reassess};
\end{tikzpicture}
\caption{Selective interactive adaptation loop with an explicit reachability gate. A foundation model can
supply the representation and prompt interface, but a separate controller governs interaction and stopping,
and adaptation cannot certify its own safety: each update is independently reassessed against clinical risk
and protected capabilities.}
\label{fig:loop}
\end{figure}

\section{A Two-Layer Memory Architecture}
\label{sec:memory}

The second capability commonly proposed for rare-case few-shot systems is learning from clinician
experience accumulated over edge cases. We treat this as a first-class part of the architecture rather than an
optional extension, but with a change of learning target, because the naive version is statistically
self-defeating.

\subsection{Why the naive target is unlearnable}

Rare cases are rare. If the object to be learned is $p(\text{mask} \mid \text{rare presentation})$, then each
class of presentation may be encountered once or twice in an institution's history, and no amount of
governance repairs a sample size of one. Worse, the presentations are heterogeneous in ways that do not
pool: an unusual postoperative anatomy and an unfamiliar reconstruction kernel share nothing at the level
of appearance.

The resolution is to change what is being estimated. Failure \emph{modes} are far fewer than diseases.
Boundary overshoot into an adjacent bright structure, topological fragmentation of a thin structure, missed
small lesions near a high-gradient interface, and spurious attachment to an artifact recur across anatomies,
modalities, and pathologies. What generalizes across rare cases is therefore not clinical knowledge but a
\emph{correction prior} over geometric and semantic failure modes: an estimate of
$p(\text{correction direction} \mid \text{failure mode}, \text{context})$. This target has a workable effective
sample size because every case that exhibits a mode contributes, regardless of its diagnosis. It also has a
natural interface to Section \ref{sec:correctable}, since the modes that admit a compact correction direction
are largely the modes that populate the correctable set.

\subsection{Layer A: the per-case episode}

The episode layer is the one formalized in Equations \eqref{eq:objective}--\eqref{eq:pnevi}. Updates are
episode-local and reversible: prototype revision, prompt optimization, adapters, or small normalization
modules, combined with source anchors, trust regions, and snapshots. Per-case test-time adaptation can
recover performance across scanner and protocol shifts \cite{karani2021tta}, and source-free domain
adaptation addresses a distinct target-cohort setting in which source images cannot be retained
\cite{bateson2022sfda}. Continual test-time adaptation and entropy minimization can accumulate
pseudo-label error or forget source capability in changing streams
\cite{wang2022cotta,niu2022eata}, although much of this evidence comes from non-medical benchmarks.
Correction-driven adaptation is established \cite{kontogianni2020continuous}, and OAIMS directly studies
online adaptation from refined masks under medical distribution shift \cite{xu2026oaims}. These are
baselines, not novelty claims. Within this layer, one intervention should correct the present mask and,
where justified, clarify the task for later slices in the same volume; nothing written here survives the
episode.

\subsection{Layer B: the cross-case experience layer}

The experience layer holds correction priors, not cases. Its unit of storage is a tuple of failure-mode
descriptor, context descriptor, correction direction, and provenance, deliberately stripped of case identity.
Its objective is not the per-episode cost in \eqref{eq:objective} but the expected reduction in future
episode cost, evaluated over a separately governed horizon with its own resource budget. Its updates are
periodic and audited rather than online. Table \ref{tab:layers} contrasts the two layers; the separation exists
so that a claim about long-run learning cannot be smuggled into an evaluation of per-case behavior, which
is the most common ambiguity in this literature.

\begin{table}[t]
\centering
\small
\caption{The two memory layers differ in unit, horizon, objective, and governance. Conflating them is what
allows an accumulating target-domain dataset to be reported as few-shot performance.}
\label{tab:layers}
\begin{tabularx}{\textwidth}{@{}l X X@{}}
\toprule
 & \textbf{Layer A: episode} & \textbf{Layer B: experience} \\
\midrule
Unit of storage & Prompts, prototypes, adapters for the current case & Failure-mode $\to$ correction-direction priors, de-identified \\
Horizon & Single case; discarded at $\tau$ & Multi-case; explicit retention period \\
Objective & Equation \eqref{eq:objective} & Expected reduction in future episode cost \\
Update timing & Online within the episode & Periodic, batched, audited \\
Write gate & Frozen monitor; rollback & Cross-reader reproducibility (Section \ref{sec:writegate}) \\
Governance & Ephemeral; no consent beyond care & Consent, provenance, deletion, protected-task tests \\
Failure mode & Within-episode drift & Institutional bias entrenchment \\
\bottomrule
\end{tabularx}
\end{table}

\subsection{The write gate: cross-reader reproducibility}
\label{sec:writegate}

Section 5.5 required distinguishing annotation error, systematic reader preference, irreducible ambiguity,
and legitimate use-case-specific contouring, but a principle is not a mechanism. We propose reproducibility
across readers as the operational criterion, because it is measurable and because it aligns the destination of
a correction with the scope of its validity.

\begin{description}[leftmargin=1.6em,itemsep=2pt]
\item[Shared write.] A correction pattern that is independently reproduced by multiple readers under
comparable context is admitted to the shared experience layer. Reproducibility is assessed on the
failure-mode descriptor and correction direction, not on pixel agreement, and requires a prespecified
minimum number of distinct readers and cases.
\item[Personal write.] A pattern reproduced by a single reader across cases, but not by others, is admitted
only to that reader's personalization profile and never influences others' outputs. This is the correct
destination for legitimate use-case-specific contouring and for systematic reader preference.
\item[No write.] A pattern observed once remains in the episode and expires with it. This is the default, and
it is where genuine annotation error is expected to land.
\item[Explicit ambiguity.] A context in which readers reproducibly \emph{disagree} is recorded as such and
routes future similar cases toward deferral or adjudication rather than toward a single prior
\cite{warfield2004staple,kohl2018probunet,baumgartner2019phiseg}.
\end{description}

The rule has a cost, which should be acknowledged: it is conservative for institutions with few readers, and
it will systematically delay learning from genuine expertise held by a single specialist. We regard this as
the correct default in a safety-critical setting, but the threshold is a parameter to be reported, and its effect
on learning rate is itself an experimental quantity.

\subsection{Governance of the experience layer}

Persistent adaptation raises privacy and governance questions because site- or clinician-specific memory
may encode identifiable or institution-specific information. Retained representations require access control,
audit trails, and deletion mechanisms. Storing correction priors rather than cases mitigates but does not
eliminate this: a sufficiently specific failure-mode descriptor can be re-identifying. The experience layer
therefore requires its own consent basis, protected-task tests before any interaction may influence later
patients, and a demonstrated transfer benefit; absent measurable transfer, forgetting is the safer default.
The scientific question is not whether a model can keep learning, but under what evidence and governance
conditions retaining an interaction is safer than discarding it.

\section{Design Principles and Research Directions}

\subsection{Calibrated recognition of correctable and non-correctable failure}

The first direction estimates residual risk at three linked levels. Voxel- or patch-level uncertainty should
localize candidate errors and support spatial querying. Structure-level uncertainty should reveal missing
components, topology violations, boundary failures, and small lesions that an average heat map can hide.
Case-level risk should govern release or deferral. Accuracy and confidence must remain separate: modern
networks are often overconfident \cite{guo2017calibration}, calibration degrades under distribution shift
\cite{ovadia2019trust}, aleatoric and epistemic components behave differently \cite{kendall2017what}, and
medical segmentation studies demonstrate both the promise and the limits of uncertainty for quality control
\cite{mehrtash2020confidence,jungo2020analyzing,zenk2025benchmarking}.

To this we add the reachability estimate $\hat{\rho}$ of Section \ref{sec:correctable} as a fourth, orthogonal
quantity. An OOD score measures atypicality, not the probability of segmentation failure
\cite{hendrycks2017baseline}; predicted mask quality measures failure, not repairability; $\hat{\rho}$
measures repairability, not correctness. An unusual case may be easy, a serious error may remain
in-distribution, and a small error may be unreachable. Conditional calibration by site, scanner, modality,
structure size, pathology, and interaction step is more informative than pooled voxel calibration. A key
negative result is scientifically meaningful: if confidence increases after feedback without improved
correctness, the signal is unsuitable for query control.

\subsection{Cost-sensitive multimodal feedback as task specification}

The second direction treats interaction modalities as semantically distinct supervision channels. Existing
systems establish points, boxes, scribbles, masks, and constrained semantic prompts
\cite{wong2024scribbleprompt,du2024segvol,kirillov2023sam}; in-context systems show that labeled
image-mask pairs can specify unseen segmentation tasks \cite{butoi2023universeg,wong2025multiverseg}.
These findings do not establish that free clinical text can explain an exception, that a reference case
reliably communicates acquisition context, or that accepting a suggested region is valid weak supervision.
Those channels should be experimental variables. Earlier work predicts sufficient annotation strength by
jointly considering accuracy and human cost \cite{jain2013sufficient}; the unresolved problem in
CD-FSMIS is response-conditioned selection \emph{among} channels, coupled to accept/query/defer and
measured in real effort.

The policy must learn where to ask, what to ask, and how to ask, and it must be evaluated against the
strongest available alternative to asking at all. A foreground point may resolve semantic identity; a short
correction may resolve geometry; a matched reference case may reveal unfamiliar appearance; and full
review may dominate when no compact answer is reliable.

\subsection{Bounded, reversible adaptation}

The third direction uses clinician feedback without converting deployment into uncontrolled online training,
and its design objective is reversibility rather than rate. A separately frozen monitor should control update
gates and rollback; operational separation requires fixed monitor parameters, calibration on data disjoint
from episode adaptation, and no updating from the same clinician response that triggered the update.
Reporting should include rollback frequency, because a system that never rolls back has either a perfect
update rule or an inert monitor, and the two are distinguishable only by measurement.

\subsection{Selective and personalized clinician-model teaming}

Clinical collaboration depends on both partners. Evidence transferred from diagnostic classification shows
that decision support can improve performance, while inaccurate advice, including advice presented as
AI-generated, can mislead clinicians \cite{tschandl2020collab,gaube2021doasaisay}. A large radiology
study found heterogeneous effects across 140 radiologists and 15 chest-radiograph tasks, with AI error a
major determinant of harm \cite{yu2024heterogeneity}. Segmentation-specific studies report gains in
contouring accuracy or time in some settings \cite{lu2021randomized}, but heterogeneous savings across
structures and institutions \cite{pang2025multicentre} and possible automation bias in prostate radiotherapy
contouring \cite{arjmandi2026automationbias}. Classification evidence is not proof for interactive
segmentation; together, these studies demand direct workflow evaluation. Uncertainty-guided selection for
preference alignment \cite{zhao2026uair} is a comparator, not a precedent.

Clinician behavior can enter the system state, but response time or correction style must not be interpreted
directly as competence, because both are confounded by case difficulty, interface familiarity, fatigue, and
prior allocation. Learning-to-defer provides a basis for routing to humans
\cite{madras2018predict,mozannar2023whopredict} and for adapting to an unseen expert from a population
\cite{tailor2024population}; it does not validate dense segmentation or transfer clinical responsibility.
Personalization should require consent, interpretable features, minimum evidence, bounded exploration,
subgroup workload audits, reversibility, and an unconditional clinician override, and its write destination is
the personal layer of Section \ref{sec:writegate}.

\section{Foundation Models as an Enabling Substrate}

The decision framework does not prescribe a backbone. Foundation models warrant separate treatment
because they can unify representation and interaction interfaces while leaving calibration, stopping, and
adaptation safety unresolved.

SAM established a broadly promptable segmentation interface \cite{kirillov2023sam}. Points, boxes, and
masks can alter an output without task-specific retraining, which makes this interface attractive for
clinician-in-the-loop inference. Yet direct retrospective medical evaluation found large variation across
tasks and strong dependence on protocol-generated prompt type \cite{mazurowski2023samstudy}. MedSAM
expands promptable segmentation through large-scale medical adaptation but primarily uses box prompting
\cite{ma2024medsam}; Medical SAM Adapter still requires task-specific parameter optimization
\cite{wu2025medsamadapter}; and FM-ABS uses active expert cross-labeling in semi-supervised model
development rather than real-time clinical inference \cite{xu2024fmabs}. Strong specialized pipelines
remain competitive baselines and should be reported as such \cite{isensee2021nnunet}. These studies
establish technical feasibility, not calibrated abstention, model-initiated queries, or prospective safety.

A foundation model should therefore be treated as a reusable representation and interaction substrate.
MAUP provides an immediate bridge: multi-center representations and uncertainty-aware prompt selection
condition a frozen SAM at inference without parameter updates \cite{zhu2025maup}. Its uncertainty score
ranks prompts; it has not established patient-level risk calibration or justified abstention. Notably, a
promptable interface also makes the reachable set of Section \ref{sec:correctable} unusually tractable,
because the admissible operator is enumerable over prompt perturbations, which is why we regard
frozen-SAM pipelines as the natural testbed for the pilot in Section \ref{sec:pilot}.

The preferred architecture is modular: a medical foundation encoder provides reusable features; episode
memory combines support examples, gated clinician-confirmed masks, and interaction history; a task head
proposes the segmentation; separate modules estimate local risk, case risk, and reachability; a controller
selects accept, query, or defer; and a constrained adapter modifies only approved components. This
separation allows each safety claim to be tested independently. A stronger backbone can improve the initial
mask, but cannot conceal a poorly calibrated controller or a drifting update rule.

\section{Experimental and Clinical Validation}

Because each interaction changes both the prediction and the remaining information state, evaluation must
address the complete clinician-model trajectory rather than only its final mask.

\subsection{A minimal pilot that can falsify the foundational claim}
\label{sec:pilot}

A purely normative framework invites the objection that it cannot be wrong. We therefore specify the
smallest study that could refute its base layer, before any clinician is recruited and without new data
collection. It uses an existing CD-FSMIS setting and a frozen promptable backbone
\cite{zhu2025maup,kirillov2023sam}, for which the admissible operator is enumerable over prompt
perturbations and the reachable set can be sampled directly.

\begin{enumerate}[leftmargin=1.8em,itemsep=3pt]
\item \textbf{Construct reachability labels.} For each held-out case, sample admissible interaction
trajectories under a fixed budget using several independent simulated-user policies, and record whether any
trajectory attains the acceptability criterion. This yields a binary reachability label without any new
annotation.
\item \textbf{Estimate $\hat{\rho}$ and test it out of domain.} Train the reachability estimator on source
domains only and evaluate discrimination and calibration at held-out target domains, stratified by shift type
and difficulty. \emph{This is the falsification point.} If $\hat{\rho}$ does not exceed chance at external
targets, or is not better than predicted mask quality alone, the base layer fails.
\item \textbf{Compare controllers at matched effort.} Evaluate the gated $\nEVI$ controller of
\eqref{eq:greedy} against voxel-entropy querying, random querying, a fixed-click heuristic, and passive
correction, holding total simulated effort equal. Report risk-coverage curves and quality-effort curves as
the two primary displays.
\item \textbf{Ablate the gate.} Run the same controller with the reachability constraint removed. If
performance is unchanged, the gate is decorative and the central formal contribution of this paper is
unsupported.
\item \textbf{Probe the response model.} Repeat under deliberately misspecified simulated responders,
including a disagreeing responder and a non-responding one, and compare the standard controller against
the pessimistic form \eqref{eq:pnevi}.
\end{enumerate}

This pilot cannot establish clinical benefit, and we do not claim otherwise; simulated users omit search time,
navigation, windowing, and cognitive switching. Its value is that four of its five steps can return a negative
result that would end the program, which is the property a normative framework most needs.

\subsection{Study domains and edge-case construction}

Full evaluation should begin with multi-institutional MRI and CT tasks that expose both controlled and
naturally occurring shifts: MRI sequence, field strength, vendor, and reconstruction; CT phase, dose, kernel,
and hardware; institution-specific protocols; and temporal drift. Task novelty should include unseen
structures and lesions. Edge-case strata should include small targets, weak boundaries, atypical
morphology, postoperative anatomy, artifact, and poor image quality. Factorial designs should separate
acquisition shift from clinical ambiguity wherever possible, so that a query policy is not rewarded for
treating every unfamiliar texture as a request for annotation.

Average external performance is insufficient. Cross-hospital studies show variable generalization and
exploitation of institution-specific signals \cite{zech2018variable}; causality-inspired domain
generalization demonstrates the need to address acquisition appearance and spurious correlation
\cite{ouyang2023causality,zhou2023dgsurvey}. Results should be stratified by site, scanner, sequence,
target size, pathology, and difficulty, with worst-group performance and low-performing quantiles reported.
Held-out institutions and prospective temporal cohorts should remain untouched until the method and
thresholds are frozen.

\subsection{Evaluation of the coupled system}

Overlap measures such as Dice remain useful for continuity but cannot define clinical success. Candidate
geometric proxies include normalized surface Dice at a clinically elicited tolerance, average symmetric
surface distance, and HD95; their relationship to editing time, inter-reader acceptability, and downstream
outcomes must be tested rather than assumed. Metrics Reloaded shows why metric choice should follow
target geometry, data hierarchy, imbalance, and decision purpose \cite{maierhein2024metrics}; clinically
applicable contouring studies illustrate tolerance-aware surface evaluation and independent expert
comparison \cite{nikolov2021clinically}.

The primary calibration target should be a patient- or structure-level probability that the complete mask
fails a prespecified, blinded acceptability criterion. On an independent calibration cohort, reliability curves,
Brier and log scores, and uncertainty intervals should be reported by deployment stratum and interaction
step. Pooled-voxel expected calibration error is secondary because it can conceal case-level miscalibration.
AUROC and AUPRC assess ranking, not threshold safety, so selective-safety reporting should include
false-accept rate, sensitivity, positive predictive value, and empirical coverage at a frozen threshold with
confidence intervals. A claim of distribution-free risk control must state its exchangeability assumptions
explicitly \cite{bates2021rcps}; otherwise ``maximum risk'' remains an empirical operating target.

A false query has no directly observed label on a single trajectory, because the safe no-query outcome is
counterfactual. Operationally it is a query whose adjudicated net expected value is non-positive: the
pre-query result was acceptable and realized or estimated benefit falls below a prespecified minimum after
measured effort. Randomized query/no-query assignment or validated offline policy evaluation is needed.
Interaction outcomes should be expressed against elapsed clinician time, corrected area or slices, interface
actions, and latency, with summaries such as time to acceptable mask, area under the quality-effort curve,
improvement per minute, and failure within a fixed budget. Decision-curve analysis offers a framework for
explicit net benefit rather than inferring clinical utility from geometry alone \cite{vickers2006dca}.

Adaptation should be evaluated after every response: immediate gain, calibration change, monotonicity,
protected-task retention, within-volume transfer, error propagation, and rollback frequency. The experience
layer of Section \ref{sec:memory} should be evaluated separately and should additionally report cross-case
transfer, write-gate admission rates, and performance over changing domain sequences. An update that
increases confidence while increasing clinically weighted error is a safety failure. Downstream endpoints
such as volumetry stability, radiotherapy contour acceptability, or derived-measurement sensitivity should
be used when the application permits.

\begin{table}[t]
\centering
\small
\caption{Evaluation follows the causal chain from local prediction to interaction, adaptation, and workflow.
The reachability row is new and is where the framework is most exposed to falsification.}
\label{tab:eval}
\begin{tabularx}{\textwidth}{@{}l X X@{}}
\toprule
\textbf{Level} & \textbf{Representative measures} & \textbf{Primary safety question} \\
\midrule
Voxel or region & Calibration, error localization, boundary distance, Dice. & Does uncertainty identify a clinically meaningful correction? \\
\addlinespace
Structure or case & NSD, ASSD, HD95, missed small targets, failure AUROC/AUPRC. & Can the system recognize a clinically unacceptable complete result? \\
\addlinespace
Reachability & Discrimination and calibration of $\hat{\rho}$; gate ablation; unreachable-case deferral rate. & Can the system tell a repairable error from an unreachable one? \\
\addlinespace
Selective policy & Risk-coverage, false accepts, counterfactual query value, frozen-threshold coverage. & Are release and escalation decisions accountable? \\
\addlinespace
Interaction & Time to acceptable mask, quality-effort AUC, improvement per minute, latency. & Does assistance save expert effort outside simulation? \\
\addlinespace
Adaptation & Immediate gain, calibration change, retention, rollback rate, within-volume transfer. & Does learning help without drift or propagated error? \\
\addlinespace
Experience layer & Cross-case transfer, write-gate admission, subgroup entrenchment audits. & Does retained memory earn its governance cost? \\
\addlinespace
Workflow & Editing time, workload, override, downstream effect, subgroup results. & Does the joint system outperform both components safely? \\
\bottomrule
\end{tabularx}
\end{table}

\subsection{From robot users to clinicians}

Simulated users are indispensable for scalable ablation but do not establish clinical interaction efficiency.
Many protocols use the ground-truth mask to place the next click and count heterogeneous modalities as one
interaction step; they therefore omit search time, navigation, windowing, and cognitive switching, and
cannot measure anchoring, fatigue, frustration, or recovery from a misleading output. Interactive
segmentation research shows that user protocols and interface design can alter conclusions
\cite{kohli2012usercentric,amrehn2019usability}. Radiotherapy workflow studies support measuring
operational and cognitive time, although small-reader studies cannot establish broad effectiveness
\cite{ramkumar2016userinteraction}.

Validation should proceed as a ladder: the pilot of Section \ref{sec:pilot}; manually collected traces
disjoint from the final human test; a reader study on representative edge cases; prospective silent
evaluation; and a limited workflow study with stopping and rollback criteria. Reader studies should use a
randomized crossover with washout or a balanced incomplete-block design. Manual delineation, review of a
raw automatic mask, a conventional interactive tool, and the proposed system should be matched where
possible for backbone, initial mask, display, training, latency, and interface. Analyses should model crossed
clinician and case effects, prespecify power, use blinded adjudication, and record quality-effort intervals,
workload, usability, override, and recovery from deliberately imperfect suggestions. A silent phase can
estimate case flow, shift prevalence, trigger rates, and potential workload; without responses it cannot
validate query utility, adaptation, anchoring, or time savings. Early reporting should follow DECIDE-AI
\cite{vasey2022decideai}, recognizing that a reporting framework is not itself evidence of safety.

\section{Falsifiable Hypotheses and Their Dependency Order}

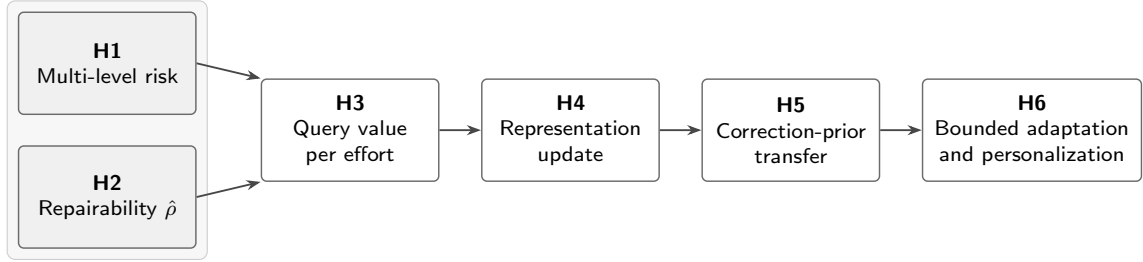
\begin{figure}[t]
\centering
\begin{tikzpicture}[
  n/.style={draw=black!58,line width=0.55pt,rounded corners=2pt,fill=white,
    align=center,font=\sffamily\scriptsize,inner sep=3.5pt,text width=2.1cm,
    minimum height=1.35cm},
  f/.style={n,fill=black!6},
  group/.style={fill=black!3,draw=black!20,line width=0.4pt,rounded corners=3pt,
    inner xsep=4pt,inner ysep=4pt},
  grouptitle/.style={font=\sffamily\fontsize{6.8}{7.5}\selectfont\bfseries,
    text=black!62,fill=white,inner xsep=2pt,inner ysep=0.5pt},
  ar/.style={-{Stealth[length=1.9mm,width=1.35mm]},draw=black!72,line width=0.7pt}
]
\node[f] (h1) {\textbf{H1}\\Multi-level risk};
\node[f,below=4mm of h1] (h2) {\textbf{H2}\\Repairability $\hat\rho$};
\begin{scope}[on background layer]
\node[group,fit=(h1)(h2)] (foundation) {};
\end{scope}
\node[grouptitle,anchor=south] at ([yshift=2pt]foundation.north)
  {FOUNDATION: NO CLINICIAN REQUIRED};
\node[n,anchor=west] (h3) at ([xshift=7mm]foundation.east)
  {\textbf{H3}\\Query value\\per effort};
\node[n,right=5.5mm of h3] (h4) {\textbf{H4}\\Representation\\update};
\node[n,right=5.5mm of h4] (h5) {\textbf{H5}\\Correction-prior\\transfer};
\node[n,text width=2.65cm,right=5.5mm of h5] (h6)
  {\textbf{H6}\\Bounded adaptation\\and personalization};
\draw[ar] (h1.east)--(h3.north west);
\draw[ar] (h2.east)--(h3.south west);
\draw[ar] (h3)--(h4);
\draw[ar] (h4)--(h5); \draw[ar] (h5)--(h6);
\end{tikzpicture}
\caption{Dependency order of the hypotheses. H1 and H2 are testable with simulated users and existing data
(Section \ref{sec:pilot}); everything downstream presupposes them. A negative result at the foundation layer
terminates the program rather than motivating a larger study.}
\label{fig:dag}
\end{figure}

The research program is useful only if its central assumptions can fail, and only if their failure has
consequences for one another. The hypotheses below are therefore ordered by dependency rather than by
importance, and Figure \ref{fig:dag} states the order explicitly. The practical implication is that the program
is sequentially abandonable: a negative result at a lower node makes the higher nodes untestable rather than
merely unsupported.

\begin{description}[leftmargin=2.6em,itemsep=4pt]
\item[H1 (foundation): multi-level risk supports action selection.] Combining local error localization,
structure quality, and case atypicality improves accept/query/defer decisions over voxel entropy alone. It is
falsified if selective risk does not improve at held-out institutions, or if patient-level calibration fails after
thresholds are frozen.

\item[H2 (foundation): repairability is estimable, and distinct from error.] The reachability score
$\hat{\rho}$ discriminates correctable from unreachable failure at external sites, beyond what predicted mask
quality and atypicality already provide, and removing the gate degrades the policy. It is falsified by
chance-level external discrimination, by subsumption under predicted mask quality, or by a null gate
ablation.

\item[H3: dynamic queries improve value per unit effort.] \emph{Conditional on H1 and H2.} At
matched clinician time, a value-guided policy reaches an acceptable mask more often than random queries,
fixed-click heuristics, or passive correction. It is falsified if the gain disappears under elapsed-time
accounting or with real users. If H1 or H2 fails, H3 is not tested, because a value estimate built on an
invalid risk or reachability signal is uninterpretable rather than merely weak.

\item[H4: feedback should update task representation.] \emph{Conditional on H3.} At
equal risk, a correction that updates support or prompt representation improves unedited slices in the same
volume more than a local overwrite. It is falsified by negligible within-volume transfer, propagated error, or
excessive protected-task loss.

\item[H5: correction priors transfer where mask priors cannot.] \emph{Conditional on
H4.} A cross-case experience layer keyed on failure modes reduces future episode cost on \emph{unseen}
rare presentations, whereas a case-similarity memory does not. It is falsified if transfer requires
presentation-level similarity, if benefit vanishes under the cross-reader write gate, or if subgroup audits
show entrenchment of institutional idiosyncrasy.

\item[H6: bounded adaptation and governed personalization.]
\emph{Conditional on H4 and H5.} Trust regions, anchors, frozen gates, and rollback retain most
target-domain gain while reducing catastrophic accumulation and forgetting; with consent and sufficient
observations, clinician-conditioned policies reduce time and unnecessary queries without increasing error or
over-reliance. It is falsified if constraints only suppress useful adaptation, if benefit is limited to robot
users, or if task allocation becomes inequitable or unexplainable.
\end{description}

\section{Levels of Evidence Required for Translation}

\textbf{Level I: benchmark and decision evidence.} A unified protocol should connect within-domain
FSMIS, CD-FSMIS, promptable foundation models, interactive refinement, and selective prediction.
Evidence at this level must establish calibration, reachability estimation, and accept/query/defer baselines,
define clinically weighted edge-case strata, and derive interaction-cost models from multiple robot policies
and initial clinician traces. An auditable controller can be evaluated before any parameter adaptation is
allowed. The pilot of Section \ref{sec:pilot} sits at the entry of this level.

\textbf{Level II: interaction and adaptation evidence.} Comparative studies should evaluate points,
scribbles, boundaries, references, and text together with episode-local prompt, prototype, or adapter updates.
Evidence must show whether feedback lowers risk for the intended reason, whether benefit transfers beyond
the edited region, and whether rollback prevents non-monotonic failure. Clinical collaboration is essential
because cost and acceptability cannot be inferred from benchmark masks.

\textbf{Level III: prospective clinician-model evidence.} Silent studies should first estimate shift, failure
prevalence, trigger rates, and potential workload without claiming interaction efficiency. Limited workflow
studies should then evaluate clinician-specific policies, inter-reader disagreement, and query utility. Any
cross-case learning through the experience layer must be a separately consented and governed protocol.
Thresholds should be frozen; model, clinician, and joint-system performance should be reported separately;
and site and subgroup analyses should be prespecified.

\section{Limitations, Governance, and Scope}

The central technical risk is that uncertainty can be unreliable precisely under severe shift. Independent
signals and conditional calibration can mitigate this, but detected shift can only block adaptation or trigger
review; it does not prove error. The reachability estimate inherits this problem in a specific form: it is
trained on simulated trajectories whose response model may not match clinical reality, so a systematic
optimism in the simulator becomes a systematic overestimate of repairability. The pessimistic formulation
of Section \ref{sec:coldstart} is a partial remedy, not a solution, and reporting sensitivity to the simulator is
mandatory rather than optional.

Feedback is another source of uncertainty. A correction can reflect ambiguity, fatigue, or idiosyncratic
preference, which is why the write gate of Section \ref{sec:writegate} is conservative by design. That
conservatism has a real cost in settings with few readers, and we do not claim to have resolved the
resulting trade-off between safety and learning rate.

Human factors create a distinct failure mode. A plausible mask may anchor a reader, while repeated
low-value queries create fatigue. Controlled studies must therefore measure override behavior, workload,
and recovery from deliberately imperfect suggestions, not only accuracy or usability. Evidence that
erroneous advice can degrade physician decisions \cite{gaube2021doasaisay,yu2024heterogeneity} makes
this a primary safety endpoint rather than a secondary one.

Finally, segmentation quality is not identical to clinical benefit. A boundary deviation may be harmless in
one application and consequential in another. Each study must define the use case, tolerance, failure cost,
and downstream endpoint before optimization. Retrospective benchmark gains, a single aggregate metric, or
the presence of a large foundation model cannot establish general clinical safety. This framework remains
normative and does not itself constitute evidence of prospective clinical effectiveness.

\section{Conclusion}

Conventional FSMIS asks how much can be learned from a few labeled images. Cross-domain FSMIS asks
whether that evidence transfers across unfamiliar acquisition domains. The clinician-in-the-loop
reformulation is often stated as a further step toward interaction and rapid adaptation, and we have argued
that this statement is true but insufficiently demanding. Interaction interfaces exist, adaptation mechanisms
exist, and neither is the binding constraint. What does not exist is a system that can say, for the case in front
of it, whether its own error is the kind a bounded intervention can repair.

The framework proposed here therefore places decidable self-assessment beneath interaction and adaptation
rather than beside them. It changes the unit of analysis from a final mask to the trajectory of a joint
clinician-model system, gates that trajectory on an explicit reachability test, treats parsimony rather than
speed as the virtue of interaction and reversibility rather than speed as the virtue of adaptation, and
separates a per-case episode from a governed experience layer whose learning target is a correction prior
over failure modes rather than a mask prior over rare diseases. Foundation models supply a reusable
interface, not a safety argument. The desired endpoint is not a model that appears universally confident, but
a system that knows when to act, when to ask, when asking cannot help, how to learn safely from an answer,
and when to stop.

\section*{Declaration of competing interest}
The author declares no known competing financial interests or personal relationships that could have
appeared to influence the work reported in this paper.

\section*{Data availability}
No new data were created or analyzed for this Perspective.

\appendix

\section{Non-myopic controller}
\label{app:bellman}

The greedy rule \eqref{eq:greedy} ignores that one answer can alter the value of later questions. A
non-myopic controller instead satisfies the constrained Bellman relation
\begin{equation}
V(s, B) = \min\Big\{ J_A(s),\; J_D(s),\; \min_{a \in \mathcal{A}_Q:\, b(a) \preceq B} \big[\lambda_C c(a) + \mathbb{E}_z V(\mathcal{T}(s,a,z),\, B - b(a))\big] \Big\},
\end{equation}
with the reachability gate applied to the third branch as in \eqref{eq:greedy}. The relation is stated for
completeness; in practice the state is high-dimensional and the expectation is taken under an estimated
response model, so approximate solution by offline policy evaluation on recorded trajectories is the realistic
route. The distinction between the greedy and non-myopic forms is empirically meaningful mainly when
channels differ sharply in bandwidth, for example when a cheap click is worth issuing only because it
determines whether an expensive reference-case query is warranted.

\section{Notation}

\begin{table}[h]
\centering
\small
\begin{tabularx}{0.94\textwidth}{@{}l X@{}}
\toprule
Symbol & Meaning \\
\midrule
$K$ & Number of static support image-mask pairs fixed before inference. \\
$B, B_t$ & Vector of hard interaction limits; remaining budget at step $t$. \\
$b(a), c(a)$ & Hard resource use and residual soft burden of action $a$. \\
$s_t, h_t, u_t$ & Episode state; interaction history; multi-level risk evidence. \\
$d_t \in \{A,Q,D\}$ & Accept, query, defer. \\
$\mathcal{T}$ & Bounded, reversible update operator. \\
$\mathcal{M}(s), \mathcal{M}^{\mathcal{C}}(s)$ & Reachable set and correctable set at state $s$. \\
$\hat{\rho}(s)$ & Estimated probability that the case is correctable at $s$. \\
$\Rset, \varepsilon_R$ & Protected reference set and retention tolerance. \\
$\mathcal{P}_a$ & Ambiguity set of response distributions used in cold start. \\
$\Lclin, C_D$ & Clinically weighted loss; cost of full expert review. \\
\bottomrule
\end{tabularx}
\end{table}

\end{document}